# A novel machine learning-based optimization algorithm (ActivO) for accelerating simulation-driven engine design

Opeoluwa Owoyele*, Pinaki Pal

Energy systems division, Argonne National Laboratory, 9700 S. Cass Ave, Lemont, IL, USA 60439

* Contact author: oowoyele@anl.gov, phone number: 630-252-2132

## Abstract

A novel design optimization approach (ActivO) that employs an ensemble of machine learning algorithms is presented. The proposed approach is a surrogate-based scheme, where the predictions of a weak leaner and a strong learner are utilized within an active learning loop. The weak learner is used to identify promising regions within the design space to explore, while the strong learner is used to determine the exact location of the optimum within promising regions. For each design iteration, exploration is done by randomly selecting evaluation points within regions where the weak learner-predicted fitness is high. The global optimum obtained by using the strong learner as a surrogate is also evaluated to enable rapid convergence once the most promising region has been identified. First, the performance of ActivO was compared against five other optimizers on a cosine mixture function with 25 local optima and one global optimum. In the second problem, the objective was to minimize indicated specific fuel consumption of a compression-ignition internal combustion (IC) engine while adhering to desired constraints associated with in-cylinder pressure and emissions. Here, the efficacy of the proposed approach is compared to that of a genetic algorithm, which is widely used within the internal combustion engine community for engine optimization, showing that ActivO reduces the number of function evaluations needed to reach the global optimum, and thereby time-to-design by 80%. Furthermore, the optimization of engine design parameters leads to savings of around 1.9% in energy consumption, while maintaining operability and acceptable pollutant emissions.

# 1. INTRODUCTION

In the quest for better fuel economy to satisfy consumer demands and lower emissions to meet increasingly stringent government regulations, many original automotive manufacturers (OEMs) are exploring alternative combustion technologies in the development of internal combustion (IC) engines. A key component in bringing such new technologies to market is optimizing engines that operate on these technologies. In doing this, engineers need to select the best design parameters for the engine out of many alternatives, where each alternative features a unique combination of the design parameters of interest. One promising and emerging technique in carrying out this selection process employs simulation-driven design optimization (SDDO). In the past decade, significant progress has been made towards developing predictive models in the context of computational fluid dynamics (CFD) simulations of IC engines. After validating these CFD models against experiments, they are coupled with optimizers to obtain promising designs for physical prototyping. SDDO offers the unique advantage of enabling researchers to optimize variables that are otherwise difficult to vary in physical engines, such as piston bowl geometry. SDDO also enables virtually testing engines in regimes that are detrimental to physical prototypes, such as conditions where the propensity for knock is high.

Nevertheless, many challenges limit the use of SDDO in IC engine-type applications. First, these engines involve minimizing a primary objective function (often the fuel consumption normalized by the power output), while taking secondary objectives into consideration (usually as constraints). In many cases, the combination of design parameters that achieves the best primary objective excessively violates the constraints of interest. These constraints are often important, and exceeding them can lead to catastrophic mechanical failure or emissions that are above acceptable limits. A second challenge is that the parameters of interest often have a moderate to high dimensionality (10 to 30 input design parameters), and these parameters affect the combustion behavior in very disparate ways. Lastly, and perhaps most importantly, the function evaluations involve running CFD simulations, which are notoriously expensive. Because of these challenges, in many cases, time-to-convergence to an optimal design can take months.

Previous studies have employed genetic algorithm (GA), design of experiments (DoE), and swarm-based techniques for optimization of IC engines. Wickman et al. [1] used a GA to perform design optimization of a combustion chamber geometry with the goal of simultaneously minimizing $NO_x$, unburned hydrocarbons, soot emissions, and fuel consumption. In other studies, genetic algorithms have been used to find the optimal fuel injection profile [2] and to co-optimize the piston bowl geometry, spray targeting, and swirl ratio levels in diesel engines [3]. More recently, Broatch et al. [4] used a genetic algorithm to optimize a compression ignition engine, while considering noise control as part of the design objective. Pei et al. [5] used a DoE approach to find the optimal piston bowl geometry, injector spray pattern, in-cylinder swirl motion, and thermal conditions with the goal of improving the fuel efficiency of a heavy-duty diesel engine while maintaining $NO_x$ emissions between 1 g/kW-hr and 1.5 g/kW-hr. Swarm-based methods have also been used in a few experimental optimization studies. Karra and Kong [6] used particle swarm optimization (PSO) and experimental techniques to find the operating conditions that minimize emissions from a multi-cylinder, turbo-charged diesel engine. Zhang et al. [7] used a variant of the artificial bee colony algorithm to experimentally minimize emissions from a diesel engine.

Optimizers based on genetic algorithms are often successful in reaching the vicinity of the global optimum, but suffer from slow convergence. The micro-genetic algorithm ($\mu$GA) [8], commonly used because of its suitability with small batch sizes, sufficiently explores the surface but displays poor local search characteristics. On the other hand, DoE is often based on linear models and is not capable of capturing non-linear interaction effects without the addition of cross-terms or higher-order terms, which can lead to a loss in accuracy. Swarm-based methods such as PSO [9] and artificial bee colony [10] operate on the principle of swarm intelligence, where members of the swarm cooperatively attempt to find the optimum on a design surface. These optimizers, however, work best with relatively large swarms and often fail to properly explore the surface with limited swarm sizes. This can lead to undesirable convergence to a local optimum when applied to engine-type applications, where the swarm size needs to be small due to the computational cost of CFD simulations.

Recently, some studies have attempted to address these limitations using various machine learning- (ML) driven approaches. Moiz et al. [11] developed a machine learning-genetic approach (ML-GA), where a Super Learner [12] was used as a surrogate model for CFD simulations, showing that the number of required simulations and the total optimization clock-time was reduced by 75% compared to a CFD-GA. Probst et al. [13] coupled a Super Learner trained using a large number of CFD simulations with various global optimizers. The study found that, in terms of successfully finding the global optimum with few function evaluations, $\mu$GA was one of the best performing optimizers. In a subsequent study, the ML-GA approach was enhanced by replacing the genetic algorithm with a grid gradient algorithm (GGA), which led to more repeatable global optima over different trials [14]. Badra et al. [15] applied the ML-GGA method to piston-bowl optimization of a light-duty gasoline compression-ignition engine, leading to savings as high as 14.7% in the indicated specific fuel consumption. More recently, Owoyele et al. [16] developed an ML-GA approach that incorporates automated hyperparameter tuning to improve the quality of the machine learning surrogate models and hence, reduce the number of required function evaluations. Kavuri and Kokjohn [17] employed Gaussian process regression to enhance a genetic algorithm approach for CFD-driven optimization. Bertram and Kong [18] performed diesel engine calibration using support vector machines to enable a particle swarm optimizer escape local minima and accelerate the calibration process. However, a major limitation of these ML approaches is that they rely on the generation of an initial DoE using CFD simulations for training ML surrogate models. This approach of sampling points from the whole design space is sub-optimal and leads to redundant inclusion of points in regions far away from the global optimum and significantly larger number of CFD simulations than actually needed, thereby increasing the overall computational cost.

In order to circumvent the shortcomings of state-of-the-art evolutionary and ML optimization approaches discussed above, the authors recently proposed a novel ensemble ML-driven Active Optimizer (ActivO) technique [19], to significantly accelerate design optimization. In this approach, two different ML surrogates called *weak* and *strong* learners, trained in an active learning loop, perform exploration and exploitation of the large design space, respectively. The weak learner successively guides the optimizer towards the promising regions within the design space and the strong learner is used to exploit those promising regions to find the global optimum. Preliminary proof-of-concept studies were performed demonstrating superior convergence rates (~ 4-5x speed up) compared to state-of-the-art optimization methods ($\mu$GA and PSO). In addition, this algorithm progresses in iterations of small number of simulations. Therefore, unlike current ML approaches, no initial DoE is required, which makes ActivO highly sample-efficient.

In this work, the authors build on their previous work [19], and present a comprehensive and robust CFD-ActivO workflow for simulation-driven design optimization. In terms of algorithmic development, the present work includes two novel contributions. On one hand, a new strategy for incorporating a dynamic balance between exploration and exploitation is presented. This is in

     

contrast to the previous study [19], where for each iteration, a fixed number of evaluation samples were obtained from the strong and weak learners. Regardless of the design problem and behavior of the learners during optimization, for a set of *N* designs, the weak learner provided *N-1* design points, while the strong learner provided only *1* design point for the next iteration. In this study, the ratio of design points provided by the two learners is allowed to dynamically change during the optimization process based on the behavior of the weak learner surface. Secondly, a robust criterion for convergence based on the weak learner is incorporated. Lastly, this is the first study where ActivO is directly coupled with CFD simulations by developing a compact workflow with an iterative two-way coupling between the optimizer and the CFD solver. The weak learner and strong learner determine the simulation parameters for a small batch of CFD simulations, and conversely, the weak learner and strong learner predictions are updated based on the data from CFD simulations. By alternating between running simulations based on ML predictions and updating ML models using information from CFD, the current approach can efficiently find promising regions of the design space, and subsequently, the global optimum. In particular, the CFD-ActivO workflow is employed for a practical IC engine optimization case and its performance is tested against state-of-the-art CFD-GA technique.

The remainder of the paper is organized as follows. Section 2 provides a detailed description of the ActivO algorithm. Then, in Section 3, two optimization test problems are described for which the performance of ActivO is analyzed. The main results are discussed in Section 4. Finally, the major conclusions are summarized in the last section.

## 2. ActivO
### 2.1. Basic ActivO algorithm

As indicated in section 1, a key challenge in CFD-based SDDO is the computational cost of running each CFD simulation. Therefore, a desired feature of optimizers for these types of tasks is fast convergence, thereby minimizing the required number of simulations. In this regard, the underlying principle of ActivO is to minimize running simulations in regions of design space where the fitness value is low, but instead, focus on regions where the predicted fitness is high. At each design iteration, to obtain the next set of design points to simulate, two types of operations are performed. In the first place, a high-bias, low-variance learner is used to generate a smooth representation of the actual surface. This learner is referred to as the weak learner. In the previous

                                                                                5

work, a 3rd order polynomial basis function model was used, while this work uses support vector regression (SVR) as the weak learner. To select design variables for the next iteration, a large number of nominees, which are randomly selected design points are generated all over the design surface. To trim these to the desired number of weak learner-generated samples for the next design iteration, a two-stage selection process is applied. During the first stage, an elitist selection strategy is employed, where points with low projected fitness values are eliminated. The fitness value corresponding to $k^{th}$ percentile (for a minimization study, or *100-k* for maximization), referred to as $\lambda_k$, is determined. Nominees with fitness values that are equal to or greater than $\lambda_k$ are retained, while those with poorer fitness values are discarded as unfit. This ensures that simulations are not needlessly performed in regions of the design space where the fitness values are projected to be undesirable. It should be noted that the fitness value used to eliminate poor designs is based on the weak learner prediction, since the actual fitness values are unknown *a priori*. In the second stage of selection, the nominees that successfully passed the first stage are screened to find out which of these are likely to provide new information about the response surface. Sampling design points that are close to points for which simulations have already been run is needless, since this does not provide significant new information. Thus, to achieve a more balanced exploration, the minimum distance between each nominee and the design points that have already been sampled, $d_{min}$, is calculated. The nominee with the highest value of $d_{min}$ is selected, since this nominee is farthest from the points that have already been sampled. Note that this nominee selection procedure is repeated until the required number of weak learner-generated samples for the next iteration are obtained. A schematic illustrating the process described above is shown in Fig. 1.

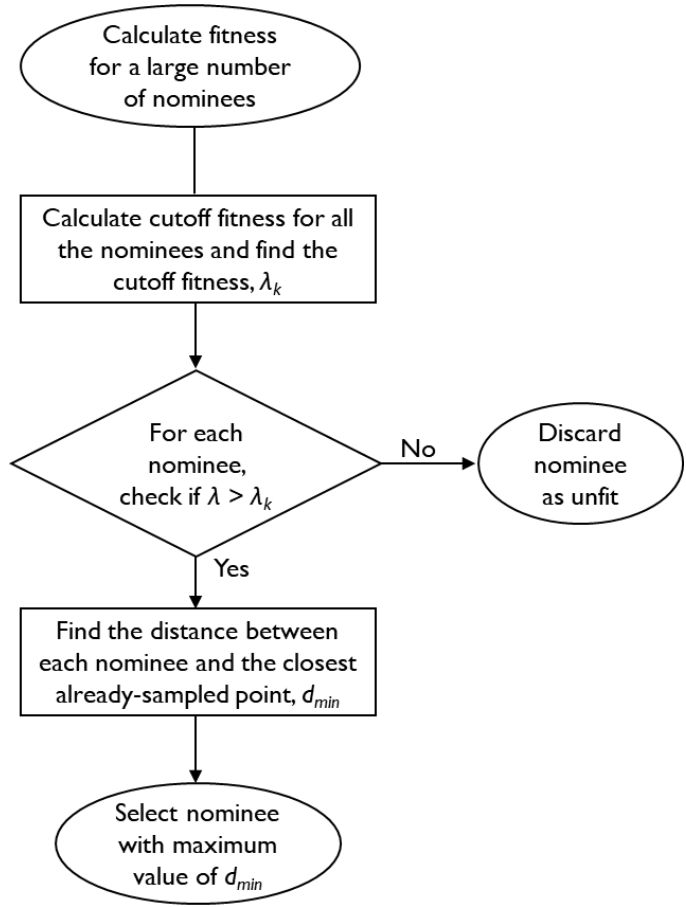

Figure 1. Flowchart showing the steps involved in obtaining new design points using the weak learner predictions

In addition to the design samples obtained from the weak learner, a low-bias, high-variance learner (called the strong learner) also contributes some samples for CFD evaluation in the next design iteration. The strong learner used in this study is a committee machine, where multiple (say *M*) artificial neural networks (ANNs) are independently trained and their predictions are averaged to get the overall prediction, $\varphi$, given by:

$$\varphi = \sum_{i=1}^{M} \varphi_i \Big/ M \qquad (1)$$

The strong learner is used in the manner of a traditional surrogate optimizer, where a differential evolution [20] algorithm is used to find the global optimum on its surface. In addition to the points obtained from the weak learner as described above, the global optimum on the strong learner surface is selected for evaluation too. Ideally, the strong learner should reproduce the finer details of the response surface with optima that are close to the best-known solution. During preliminary testing, it was observed that using a single ANN as the strong learner often led to false optima in regions far away from the best-known solution, due to overfitting that arises from having sparse sampling in non-promising regions. Averaging the predictions from multiple ANNs as done in Eq. (1) mitigates this undesired effect of sparse sampling. The neural network used as the strong learner in this study consisted of two layers with 10 neurons in each layer. The ANN neural network weights were optimized to predict the target variables using Adam optimizer [21] with a constant learning rate of 0.05. Early stopping was employed to terminate training when the training errors stop improving. For the weak learner (SVR), the cost, which determines how much to penalize points that violate the margin, was set to 16. The kernel parameter was set to the inverse of the number of features, while υ, which is a parameter that controls the number of support vectors was set to 0.5.

Overall, the weak and strong learners are meant to complement each other. Since the strong learner usually produces samples close to the best known optimum for CFD evaluation, it does not sufficiently explore the surface. Therefore, using the strong learner alone by taking its predicted optima would often lead to the optimizer getting stuck in local optima, if the surface is multimodal. On the other hand, the weak learner can provide information about areas of the design space where the fitness is generally high, but it does not provide local information. Therefore, only taking random samples from the regions with high weak learner-predicted fitness would lead to slow convergence. However, when both learners are combined, the weak learner guides the strong learner towards promising regions, while the strong learner finds the exact location of the global optimum once the promising region has been correctly identified.

**2.2. Dynamic exploration-exploitation balance and convergence criteria**

In this subsection, a method for adjusting the ratio of utilization between the two learners and the criteria for convergence is introduced. Before the design iterations begin, a large number of monitor points, randomly selected over the entire design space, are defined. The weak learner predictions at these monitor points are used to adjust the ratio of the number of points obtained from the weak learner, $p$, to those obtained from the strong learner, $N-p$, at each design iteration. First, for a maximization problem, a promising region is defined as the region of the design space where the weak learner-predicted fitness value is within the $k^{th}$ percentile (or below the $(100 - k)^{th}$ percentile, for a minimization problem). For a set of $M$ monitor points within the promising region, $\Phi = (\phi_1, \phi_2, \phi_3, \ldots \phi_M)$, the maximum percentage relative change at iteration, $i$, is defined as:

$$\omega^i = \max \left| 100 \times \frac{\Phi^i - \Phi^{i-1}}{\Phi^{i-1}} \right| \% \qquad (2)$$

The value of $\omega$ at each design iteration indicates how much exploration is taking place. A high value of $\omega$ indicates considerable changes to the surface of the weak learner within the promising region as a result of high explorative activity. This typically occurs during earlier iterations, when ground-truth information regarding the surface is very limited. As more simulation data is generated, the information gained from new simulations does not radically change the weak learner representation of the surface and $\omega$ gradually reduces. Three phases of optimization are defined, namely, extensive exploration (phase 1), preliminary exploitation (phase 2), and intensive exploitation (phase 3). Phase 1 corresponds to increasing values of $\omega$, and a high degree of uncertainty regarding the promising region. Hence, only design points from the weak learner are selected for sampling, since the strong learner is highly unlikely to find the global optimum at this stage. In stages 2 and 3, there is a 75:25 and 50:50 split between the number of weak learner and strong learner points, respectively. The optimizer moves between these three phases based on the changes in $\omega$. Each time ω increases, the optimization scheme moves down in the intensity of exploitation (from stage 3 to 2, or from stage 2 to 1). On the other hand, if there is a decrease in $\omega$, it moves in the opposite direction. The only exception to this rule occurs when $\omega$ < 5%. In this case, changes to the weak learner surface are counted as noise and the intensity of exploration is not increased even when $\omega$ increases. This is done because $\omega$ never actually fully decays to *zero*, but instead, always experiences slight variations even in the presence of insignificant changes in the ground-truth information from CFD. Overall, the goal is to utilize the weak learner more in the early stages of the optimization process, while increasingly relying on the strong learner as the global optimum is approached. The dynamic adjustment of the balance between exploration and exploitation is summarized in table 1.

                                                                  

Table 1. Summary of dynamic combination of weak learner and strong learner

| Optimization phase | $p$:$N$-$p$ | Action |
|---|---|---|
| Extensive exploration | 100:0 | If $\omega$ increases, do nothing. If $\omega$ decreases, move to phase 2. |
| Preliminary exploitation | 75:25 | If $\omega$ increases, move to phase 1. If $\omega$ decreases, move to phase 3. |
| Intensive exploitation | 50:50 | If $\omega$ increases and new $\omega$ is above 5%, move to phase 2. If decreases or increases to a value less than 5%, do nothing. |

The behavior of $\omega$ is also used to assess convergence. The two criteria used to define convergence are static exploration and stagnant exploitation. Static exploration refers to a situation where the weak learner surface is not changing with information coming from new samples. The promising region remains reasonably fixed, and the weak learner predictions within this region are also steady. As in the dynamic adjustment of $p$, this can also be determined by $\omega$. Stagnant exploitation refers to an unimproved fitness even with intensive exploitation after a predetermined number of design iterations. As opposed to requiring fully stagnant exploitation (no improvement to the fitness) for convergence, for expensive simulations, it may be expedient to define this based on a tolerance parameter, $\varepsilon$, instead. This is because, if full stagnation is required, insignificant improvements to the fitness can keep the simulations running for significantly longer than is practically necessary. Therefore, in this study, stagnation is considered to have been reached when the improvement in fitness is less than $\varepsilon$. Overall, convergence is assumed if $\omega$ remains below 5% and the improvement in the best fitness is less than $\varepsilon$ for 5 successive iterations. A schematic depicting the overall ActivO algorithm is shown in Fig. 2. The algorithm was implemented in Python 3.6, employing Tensorflow [22] and Scikit-learn [23] libraries. In this work, ActivO is applied to two optimization test problems described in the next section.

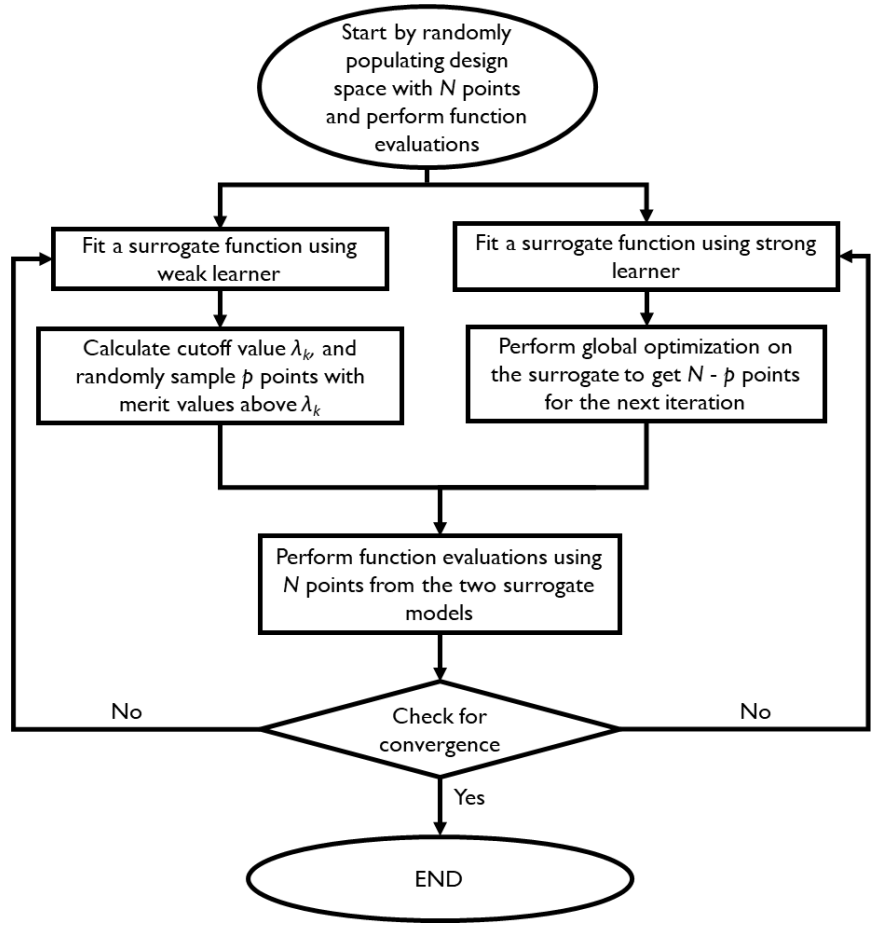

Figure 2. Flowchart depicting the ActivO Algorithm

## 3. Optimization test problems
### 3.1. Cosine mixture function

The first test case is a two-dimensional cosine mixture function [24], which is a maximization problem consisting of 25 local maxima and one global maximum. As with surfaces encountered in IC engine optimization, the cosine mixture surface, while multi-modal, still has a discernable global trend. Furthermore, using this problem as a test function allows for easy visualization and illustration of the complementary nature of the relationship between the strong and weak learners. The fitness, $z$, is defined as a function of design parameters, $x$ and $y$, as,

$$z = 0.1 \left(\cos 5\pi x + \cos 5\pi y\right) - (x^2 + y^2) \tag{3}$$

In the 2-dimensional cosine mixture function, $x \in [-1, 1]$, $y \in [-1, 1]$, and the global maximum is 0.2, occurring at $x = y = 0$. A surface representation of the function is shown below.

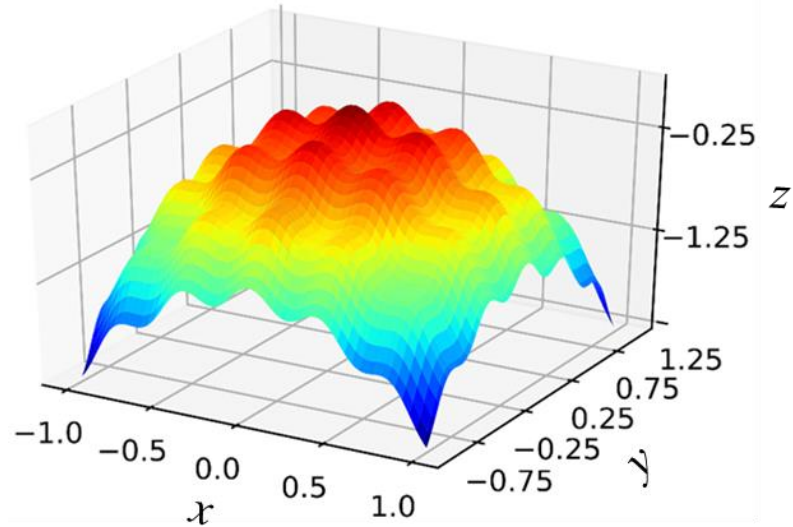

Figure 3. Surface plot of 2-dimensional cosine mixture function

## 3.2. Compression Ignition engine optimization

ActivO was also applied to the optimization of a turbocharged Cummins heavy-duty engine with a charged air cooler and exhaust gas recirculation (EGR) [25]. The engine is a compression ignition engine operating on a gasoline-like fuel at medium-load conditions. A RON70 primary reference fuel (PRF) blend comprising of 70% iso-octane and 30% n-heptane by mass was employed as the surrogate for gasoline-like fuel in the CFD simulations. A summary of the engine operating conditions is provided in table 1.

Table 1. Summary of engine operating conditions [25].

| Engine model | Cummins ISX15 |
| --- | --- |
| Displacement | 0.0149 m$^3$ |
| Bore | 0.137 m |
| Stroke | 0.169 m |
| Connecting rod | 0.262 m |
| Compression ratio | 17.3:1 |
| Engine speed | 1375 rpm |
| Intake valve closing | -137 °CA after top dead center (ATDC) |
| Exhaust valve opening | 148 °CA ATDC |
| Injection duration | 15.58 °CA |
| Mass of fuel injected | $1.245 \times 10^{-4}$ kg/cycle/cylinder |
| Fuel injection temperature | 360 K |
| Global equivalence ratio | 0.57 |

A commercial CFD code that features gradient-based adaptive mesh refinement, CONVERGE [26], was used for the CFD simulations in this study. Chemical kinetics was solved using a reduced mechanism containing 48 species and 152 chemical reactions [27]. Using acetylene ($C_2H_2$) as the precursor, the Hiroyasu soot [28] and the Nagle and Strickland-Constable [29] model was used to represent the evolution of soot. Spray breakup was modeled using the Kevin-Helmholtz Rayleigh-Taylor model [30], while the collision between spray parcels was modeled using the Schmidt and Rutland model [31]. A blob injection model, introduced by Reitz and Diwakar [32] was used to initialize the liquid droplets. A Reynolds-averaged Navier-Stokes re-normalized group [33] model (RANS RNG $k$-$\varepsilon$) was employed to model in-cylinder turbulence. To reduce computational cost, the computational geometry was modeled as a sector mesh with periodic boundary conditions in the azimuthal direction. For complete details of the CFD model setup and validation against experimental data, the readers are referred to Pal et al. [25].

In this study, the goal is to find the combination of control parameters that attain the lowest indicated specific fuel consumption (ISFC), while satisfying desired constraints. Five of these control parameters, the number of nozzle holes, nozzle inclusion angle, total nozzle area, injection pressure and start of injection timing, are related to injector design and injection strategies. The remaining variables, temperature and pressure at intake valve closing (IVC), EGR fraction and swirl ratio, are related to initial thermodynamic and flow conditions. The baseline design parameters and their corresponding ranges are shown in Table 3.

Table 3. Summary of engine operating conditions

| Parameter | Description (units) | Minimum value | Maximum value |
|---|---|---|---|
| nNoz | Number of nozzle holes | 8 | 10 |

| TNA | Total nozzle area | 1 | 1.3 |
| Pinj | Injection pressure (Pa) | $1.4 \times 10^8$ | $1.8 \times 10^8$ |
| SOI | Start of injection timing (°CA ATDC) | -11 | -7 |
| Nang | Nozzle inclusion angle (degrees) | 145 | 166 |
| EGR | EGR fraction | 0.35 | 0.5 |
| Tivc | IVC temperature (K) | 323 | 373 |
| Pinv | IVC pressure (bar) | 2.0 | 2.3 |
| SR | Swirl ratio | -2.4 | -1 |

The constraints considered in this study are related to maximum in-cylinder pressure (PMAX), maximum pressure rise rate (MPRR), as well as the mass of soot and NOx normalized by the power output ($M_{\text{soot}}$ and $M_{\text{NOx}}$, respectively).

The fitness, in this case, is then defined as follows based on the previous study by Moiz et al. [11]:

$$F = 100 * \left\{ \frac{160}{ISFC} - 100 * f(PMAX) - 10 * f(MPRR) - f(M_{\text{soot}}) - f(M_{\text{NOx}}) \right\}$$

(6)

where,

$$f(PMAX) = \begin{cases} \frac{PMAX}{220} - 1, & PMAX > 220 \ bar \\ 0, & PMAX \leq 220 \ bar \end{cases}$$

$$f(MPRR) = \begin{cases} \frac{MPRR}{15} - 1, & MPRR > 15 \ bar/CA \\ 0, & MPRR \leq 15 \ bar/CA \end{cases}$$

$$f(M_{\text{soot}}) = \begin{cases} \frac{M_{\text{soot}}}{0.0268} - 1, & M_{\text{soot}} > 0.0268 \ g/kWh \\ 0, & M_{\text{soot}} \leq 0.0268 \ g/kWh \end{cases}$$

$$f(M_{\text{NOx}}) = \begin{cases} \frac{M_{\text{NOx}}}{1.34} - 1, & M_{\text{NOx}} > 1.34 \ g/kWh \\ 0, & M_{\text{NOx}} \leq 1.34 \ g/kWh \end{cases}$$

Based on the above equation, designs where the constraints are above acceptable limits are penalized, to encourage the optimizer to find designs that do not violate these constraints. The constraints are set on PMAX and MPRR due to mechanical limits and to reduce knock propensity, respectively. Constraints on the mass of soot and NOx are added to satisfy emissions constraints.

## 4. Results and discussion

### 4.1. Cosine mixture optimization

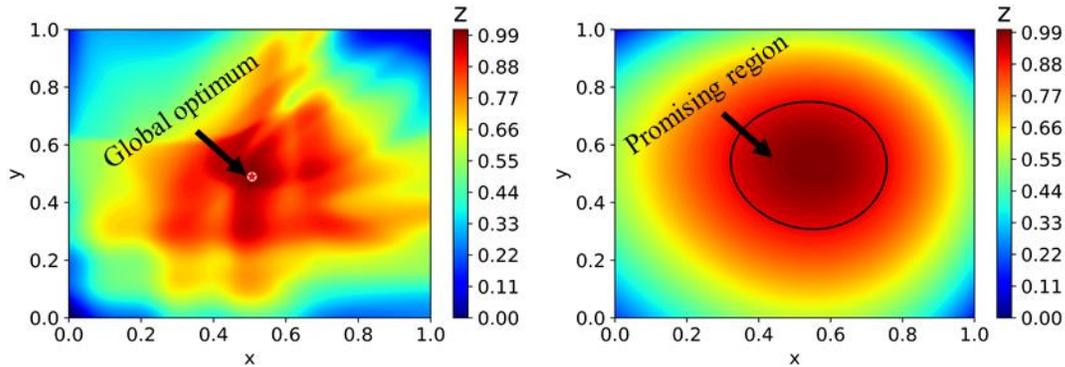

Figure 4. Strong learner (left) and weak learner (right) surfaces after convergence to the global optimum.

First, the weak learner and strong learner representations of the surface at convergence are shown in Fig. 4. As discussed in section 2, it can be seen that the strong learner surface has more details compared to the weak learner surface. The weak learner correctly finds that in general, points that are around $x = y = 0$ have higher fitness values, and therefore it identifies this region as promising while ignoring local optima. It should be noted that in many trials, the weak learner starts from the wrong location and the promising region floats on the surface until the correct region is identified. On the other hand, the strong learner surface contains more detail which enables it to find the exact location of the global optimum, after the weak learner finds the promising region.

The performance of ActivO is compared with those of five other global optimizers. First, ActivO is compared to a micro-genetic algorithm ($\mu$GA), which operates by propagating traits of individuals with higher fitness while discarding individuals with lower fitness. In contrast to other genetic algorithms, $\mu$GA involves multiple restarts by keeping the best individual and reinitializing others when the diversity in the population falls below a given threshold. This prevents premature convergence to local optima when used with small population sizes. $\mu$GA is chosen for comparison because it is the most commonly used optimizer for engine optimization as a result of its ability to work with small population sizes as is typically needed for expensive CFD simulations. For this test case, micro-convergence was assumed to occur for $\mu$GA when the variation of the binary traits within a population was less than 5%. The second optimizer was particle swarm optimization (PSO), which is a metaheuristic optimization technique that operates by updating the position and velocity of a swarm population over successive design iterations. The inertia weight of PSO, which controls the exploration-exploitation balance was set at 0.8. The third optimizer was differential evolution [20] (DE), which is a gradient-free metaheuristic algorithm that works by combining existing candidates to create new candidates while preserving the candidate with the best fitness. The fourth, genetic algorithm optimization using derivatives [34, 35] (GENOUD), approaches optimization by combining genetic algorithm methods with derivative-based quasi-newton methods. Finally, we also compare ActivO with basin hopping [36] (BH) which is a global optimizer that involves random perturbation of coordinates, local minimization, and rejection or acceptance of new coordinates. For all cases, 25 independent trials were performed to obtain meaningful projections of the performance of the optimizers.

For every optimizer, all trials involved randomly initializing the initial population of designs by drawing from a uniform distribution and utilizing a maximum of 1000 function evaluations to find the global optimum. Also, except for BH which is not a population-based strategy, 5 function evaluations per design iteration were performed for all optimizers. Figure 5 shows the evolution of the best fitness, averaged across all trials, as a function of the number of function evaluations. It can be seen that ActivO converges to the global optimum faster than all the other optimizers. In particular PSO, DE, and GENOUD converge prematurely and fail to reach the global optimum. The average maximum fitness values reached by ActivO, $\mu$GA, PSO, GENOUD, DE, and BH were 2.0, 2.0, 0.17, 0.076, 0.707, and 0.194, respectively. ActivO also exhibits the least variation of performance between different trials. After 100 function evaluations, the standard deviation in the maximum fitness across the 25 trials falls below $1\times10^{-3}$ for ActivO, compared to 555 function evaluations for $\mu$GA. In contrast, the standard deviations of PSO, DE, GENOUD, and BH remain relatively high at 0.059, 0.14, 0.16, and 0.03, even after 1000 function evaluations.

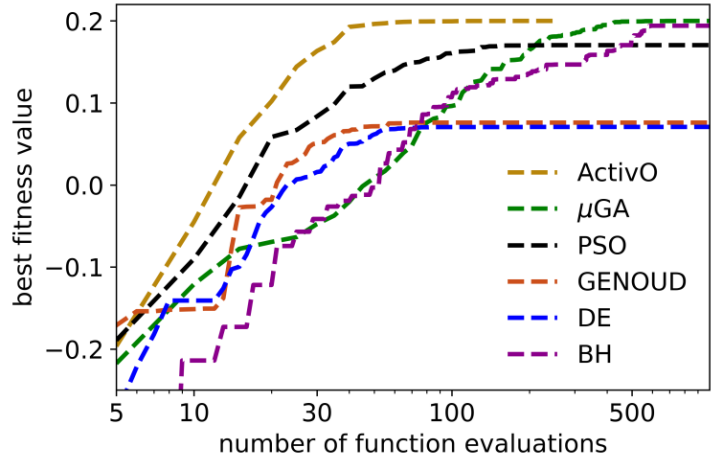

Figure 5. The evolution of the best fitness value for the cosine mixture problem using different global optimizers.

In Fig. 6, the number of function evaluations needed to reach some specified fitness values is shown. Missing bars in the figure indicate that an optimizer failed to reach the threshold fitness value of interest. From the figure, it can be observed that the average maximum merit from only ActivO and $\mu$GA successfully reach 0.198. In the figure, it can be observed that ActivO is about *5 to 7* times faster than $\mu$GA in reaching the thresholds. PSO and BH succeed in reaching 0.15, while DE and GENOUD only reach the 0.1 fitness threshold. One other point of comparison between the algorithms is shown in Fig. 7, where the probability of locating the global optimum is assessed. This probability was obtained by finding the proportion of trials that had reached the global optimum after a specific number of function evaluations. For instance, 10 trials out of the 25 finding the global optimum would amount to a probability of 0.4. A cumulative histogram was constructed based on this fraction to represent the probability of convergence. For all the cases, a tolerance of 0.002 was used, and the optimizers were assumed to have reached the global optimum when their maximum fitness value, $z_{max}$ was above 0.198. Based on these metrics, only ActivO and $\mu$GA have a 100% success rate of reaching the global optimum after 1000 iterations. For PSO, DE, BH, and GENOUD, this success rate is 80%, 4%, 96%, and 44%. It can be observed that $\mu$GA achieves a convergence probability of 1.0 after 662 function evaluations while ActivO achieves this in just 100 function calls. Therefore, for other algorithms besides ActivO and micro-GA, there is a risk that the optimizer would not find the global optimum within 1000 function evaluations. While performance may improve for some of the algorithms if more function evaluations are allowed, such a large number of function calls would require significant computing resources if coupled with a CFD solver. The failure of some of the algorithms to reach the design optimum over all trials is perhaps due to the small population used in each design iteration. Evidently, DE and GENOUD fail to adequately explore the surface using such a small population size. On the other hand, for PSO, tuning the inertia weight to achieve a good balance between exploration and exploitation becomes a challenge with such a small population size even for a two-dimensional variable space. For engine design optimization, the poor performances of these optimizers may be even more pronounced. For instance, in a recent study [19] by the authors, PSO, $\mu$GA, and ActivO were tested on an engine-like surface obtained by training a surrogate using a large number of engine CFD simulations. The study showed that the PSO performed worse than both ActivO and $\mu$GA in successfully finding the global optimum consistently using a small swarm size. Thus, relative to the other algorithms, ActivO and $\mu$GA demonstrate a better capability of working efficiently with small batches of simulations as are typically required for compute-intensive CFD-driven design optimization. Therefore, these two algorithms will be further compared with each other for the IC engine optimization performed in the next subsection.

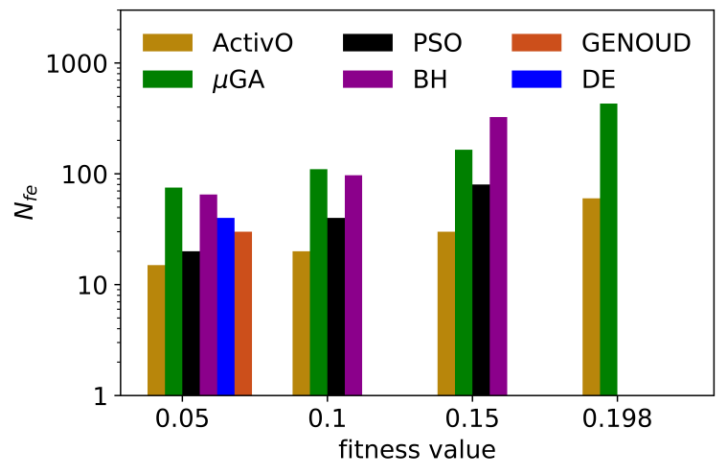

Figure 6. Number of function evaluations ($N_{fe}$) required to reach various fitness values.

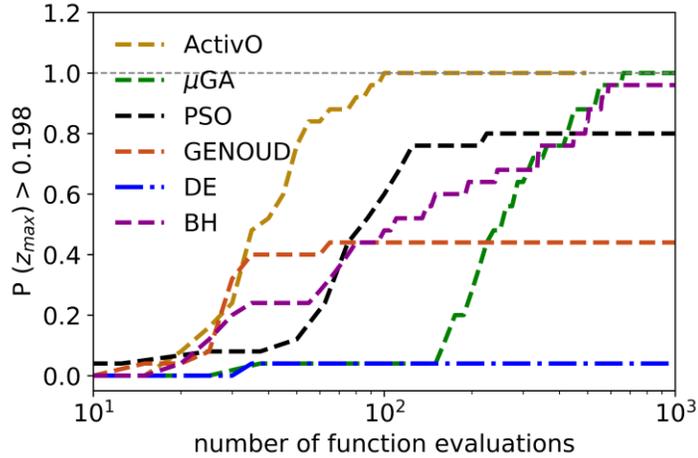

Figure 7. Improvement in the probability of convergence with increasing function evaluations.

## 4.2. IC engine optimization

For the IC engine optimization problem, as in the cosine mixture function, ActivO was compared to a micro-genetic algorithm. However, due to the computational costs of running simulations, limited runs were carried out in this case. 5 trials were run for ActivO, and this was compared with the results obtained by applying $\mu$GA to the same problem from a previous study [11] using CONVERGE's inbuilt $\mu$GA algorithm within the CONGO utility [26]. A population size of 8 individuals per generation was used. Micro-convergence was assumed for $\mu$GA when the variation in the population based on the real traits was less than 3%. After every micro-convergence event, the best individual was carried to the next generation while all others were reinitialized. For ActivO, 8 CFD simulations per design iteration were performed, analogous to the $\mu$GA algorithm.

For each trial, the simulations were initialized using 8 random combinations of the parameters as defined in section 3. The ML models were retrained after every design iteration with the information from all the previous iterations. After each batch of simulations, an in-house Python script was used to postprocess the results and extract relevant metrics related to *ISFC*, *PMAX*, *MPRR*, $M_{soot}$, and $M_{Nox}$. Based on the fitness values calculated from these quantities and the weak learner and strong learner surfaces, the design samples for the next batch of CFD simulations were determined. This iterative process was continued until convergence. It was expected, that based on the weak learner surface, the majority of the simulations would be located in regions of the design space with projected high fitness values. Also, based on the value of $\omega$, the number of strong learner points were varied. The number of design points contributed by the strong learner, in this case, were 4, 2, and 0 during intensive exploitation, preliminary exploitation and extensive exploration, respectively. Figure 8 shows the evolution of $\omega$ for all the 5 trials with the number of function evaluations. Here we see that $\omega$ (shown on a log-scale) decreases as the number of simulations increase. It starts out at values between 10 and 100, and as weak learner representation of the surface becomes more steady, the profile flattens out.

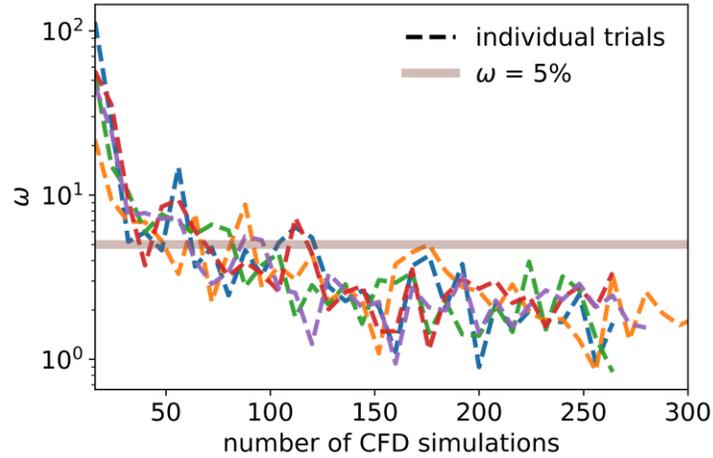

Figure 8. Decrease in $\omega$ as ActivO optimization progresses

The evolution of best fitness values from the five ActivO runs are shown in Fig. 9 versus the number of CFD simulations run. Due to the non-deterministic nature of ActivO, the figure shows a significant variation for the different runs. To quantify the performance of ActivO, the number of CFD simulations it takes to reach a fitness value of 104 was assessed (see Table 4). This value was chosen based on a number of previous studies, where it was found that the global optimum was close to a fitness of 104.0. Out of the five trials, the best case required 40 CFD simulations (5 design iterations) to reach 104.0, the worst case required 136 simulations (17 iterations) and the median case required 88 simulations (11 iterations).

Table 4. Summary of ActivO's performance on the engine optimization problem

|  | Number of design iterations to reach a fitness of 104 | Number of design iterations to achieve convergence |
| --- | --- | --- |
| Trial 1 | 5 | 12 |
| Trial 2 | 16 | 23 |
| Trial 3 | 9 | 15 |
| Trial 4 | 17 | 22 |
| Trial 5 | 11 | 17 |
| Median | 11 | 17 |
| Mean | 11.6 | 17.8 |

The optimum design parameters obtained from the 5 ActivO trials are shown in Table 5. On the other hand, the points of convergence for the ActivO trials are also depicted in Fig. 9 using the circle symbols. To assess convergence, an improvement tolerance (as discussed in section 2.2) of 0.1 was set, so that the algorithm was considered to have converged if the improvement to the fitness was less than 0.1 after 5 iterations. This is helpful in cases with expensive function evaluations, because in some of the tests carried out, the algorithm needlessly ran significantly longer due to improvements to the fitness value that was much smaller than the uncertainties associated with the simulations. In this case, the median performing case took 136 simulations to reach convergence. The best and worst cases required 96 simulations and 184 simulations, respectively (as noted in Table 4).

Table 5. Optimum design parameters

| Design Parameter | Optimum parameters found | | | | |
|---|---|---|---|---|---|
| | Trial 1 | Trial 2 | Trial 3 | Trial 4 | Trial 5 |
| nNoz | 10 | 10 | 10 | 10 | 10 |
| TNA | 1.07 | 1.09 | 1.12 | 1.18 | 1.09 |
| Pinj | 1481 | 1406 | 1415 | 1421 | 1413 |
| SOI | -10.69 | -9.66 | -10.5 | -10.3 | -10.3 |
| NozzleAngle | 157 | 155 | 156 | 157 | 157 |
| EGR | 0.45 | 0.44 | 0.45 | 0.45 | 0.45 |
| Tivc | 323 | 323 | 323 | 323 | 323 |
| Pivc | 2.3 | 2.3 | 2.3 | 2.3 | 2.3 |
| SR | -1.84 | -1.93 | -1.98 | -2.0 | -2.0 |

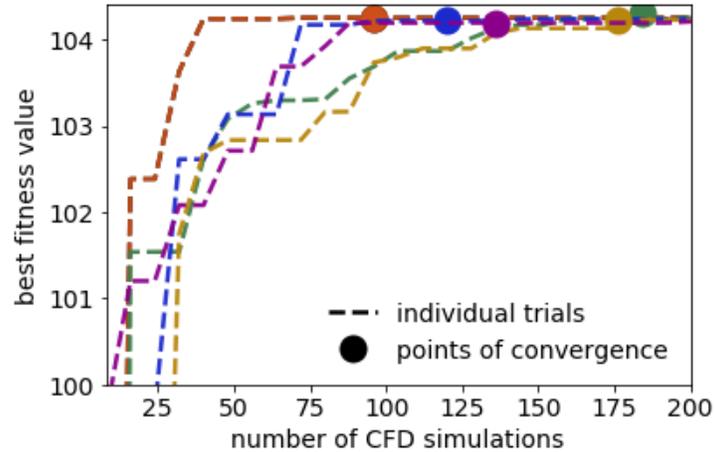

Figure 9. The evolution of the best fitness value for the IC engine optimization problem obtained from 5 trials using ActivO.

Figure 10 shows the merit values for simulation cases sampled at each design iteration for one selected trial. In addition, the points evaluated based on the strong learner are distinguished from those based on the weak learner. It can be seen that initially, all the points evaluated are based on the weak learner. At this stage, there are only a few points for training the learners, and therefore the algorithm operates by extensively exploring the design space. As the optimization progresses, the fraction of points from the strong learner increases. The plot also indicates that the strong and weak learners help exploit and explore the design space, respectively. This is evident in that the points from the strong learner seem to be clustered around the best-known merit, while the weak learner seems to select points that are far away from the merit while exploring. However, the mean merit values of the weak learner-generated points appear to increase as the optimization progresses, suggesting that the weak learner is exploring more promising regions.

                                                                                       22

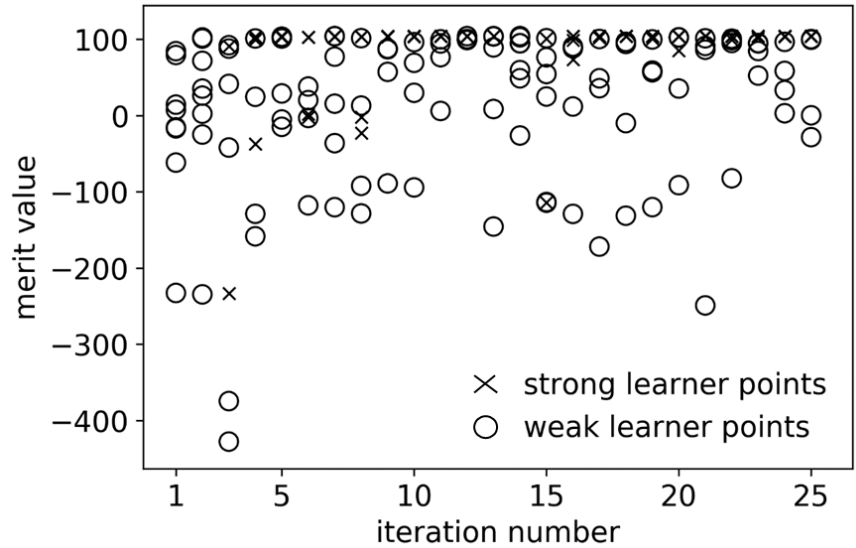

Figure 10. Population-based figure showing merit values of the points selected from the weak and strong learners

A comparison of the mean best fitness value obtained from the 5 ActivO trials and $\mu$GA is shown in Fig. 11. In contrast to ActivO's performance as discussed above, $\mu$GA converges after 784 simulations. To reach a fitness value of 104, $\mu$GA takes 58 design iterations (464 simulations), compared to 11 iterations (88 simulations) for ActivO. While only one trial is shown for $\mu$GA in this work due to the large duration of time it takes to converge (typically a few months), in a previous study [19], the authors tested the $\mu$GA on a surrogate surface obtained by training a Super Learner [12] model based on 2000 CFD simulations. Using the trained Super Learner model as a surrogate for CFD simulations and running multiple trials of $\mu$GA it was shown that on average, $\mu$GA required 80 iterations (640 simulations) to reach the vicinity of the global optimum. Since the surrogate was developed based on the same engine setup as used in this study and has been shown to lead to similar global optima in another study [11], it can be assumed that the Super Learner surface is similar to the actual engine optimization surface. It may, therefore, be reasonable to assume that the $\mu$GA performance on that surface is indicative of how it will perform on the actual surface, on average.

Table 6 compares the performances of different optimization methods for the present engine design problem, in terms of the quality of the optimum design and computational resources required. In the table, ML-GA represents the use of a Super Learner surrogate and a genetic algorithm for optimization based on Moiz et al. [11]. This is in contrast to $\mu$GA where the optimizer is coupled with CFD directly. It can be seen that ActivO achieves the highest improvement in merit value over the baseline case. The average lowest value of ISFC reached by ActivO over the 5 trials is 153.6 g/KWh, while the ISFC of the baseline design is 156.5 g/KWh. By performing design optimization with ActivO, fuel consumption savings of 2.9 g/KWh is achieved, which translates to a reduction of around 1.9%. The savings obtained using ML-GA and $\mu$GA are smaller at 2.53 g/KWh and 2.65 g/KWh, respectively. On the other hand, both ML-GA and $\mu$GA require significantly higher number of simulations (by factors of approximately 2 and 5) to converge compared to ActivO. Therefore, ActivO leads to a more efficient engine design while requiring significantly fewer CFD simulations to converge. This demonstrates the capability of ActivO to enable significant computational cost and time savings (from months to a few days) for complex design optimization campaigns. For the CFD optimization study, on average each simulation took 8.3 hours on 18 Intel Xeon E5-2695 cores, which translates to a computational cost of 1,200 core-hours per design iteration. In terms of the resources required to reach convergence, this leads to a total computational cost of 21,360 core-hours for ActivO compared to 96,000 core-hours for $\mu$GA and 39,265 core-hours for ML-GA.

In this study, ActivO was used for optimizing engine design parameters relating to fuel injection, thermodynamic conditions, and in-cylinder flow while considering a single speed-load point. However, the algorithm can be readily extended to more comprehensive and practical engine design optimization involving piston bowl design geometry and multiple speed-load conditions, which will be demonstrated in future works.

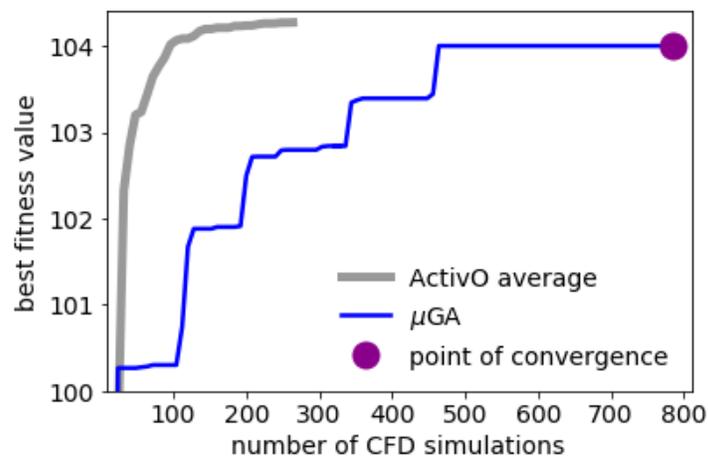

Figure 11. Comparison of the best fitness from ActivO and $\mu$GA for the engine optimization problem.

Table 6. Comparsion of the performances of ML-GA, $\mu$GA and ActivO

|  | Baseline | ML-GA | $\mu$GA | ActivO |
|---|---|---|---|---|
| Optimum ISFC (g/KWh) | 156.53 | 153.97 | 153.85 | 153.6 |
| Optimum merit | 102.2 | 103.91 | 104.0 | 104.14 |
| Number of simulations | - | 250 | 640 | 136 |

## 5. Conclusions

In this paper, a new machine learning-driven optimization scheme for accelerating design optimization was introduced. The proposed formulation is a surrogate-based approach that uses a weak learner for exploration and a strong learner for exploitation. A mechanism for dynamically adjusting the balance between exploration and exploitation, as well as a method for assessing convergence was also presented. As a first step, the proposed method was applied to optimize a 2-dimensional cosine mixture function. Its performance was compared to that of 5 different optimizers which include a micro-genetic ($\mu$GA) algorithm, particle swarm optimization, differential evolution, genetic optimization using derivatives, and basin hopping. It was shown that on average, ActivO was able to achieve better fitness values with the same number of function evaluations, compared to the other algorithms. Furthermore, it was found that ActivO produced a better confidence in finding the design optimum. Among all the algorithms tested, only ActivO and $\mu$GA successfully found the design optimum for all 25 trials. The probability of finding the optimum was projected to be 1.0 after 100 function evaluations for ActivO. Similarly, $\mu$GA was also successful in finding the optimum in all cases but took 662 function evaluations for all the trials to reach the optimum.

Subsequently, ActivO was applied to an internal combustion engine optimization case that involves finding the optimal combination of 9 control parameters to improve fuel economy while satisfying necessary constraints. In this case, 5 trials were run, and the convergence properties were presented and discussed. By performing design optimization, the indicated specific fuel consumption was reduced by around 1.9%, from 156.5 g/KWh to 153.6 g/KWh. The results were compared to those obtained from $\mu$GA in a previous study, showing faster convergence speeds for all the trials. On average, ActivO was 5 times faster than $\mu$GA in reaching a fitness value of 104 or greater. ActivO was also shown to be about 5.5 times faster, compared to $\mu$GA in attaining convergence. In general, this has the potential to provide huge savings in terms of total simulation clocktime. As an example, for a case where a simulation is assumed to take 1-2 days to complete, $\mu$GA takes 3-6 months to achieve convergence, whereas ActivO takes only 1.5 weeks to 3 weeks. In future studies, ActivO will be extended to other optimization studies that reflect more comprehensive and practical engine optimization endeavors, such as the inclusion of piston bowl geometry and multiple speed-load conditions.

**ACKNOWLEDGEMENTS**

The submitted manuscript has been created by UChicago Argonne, LLC, Operator of Argonne National Laboratory (Argonne). The U.S. Government retains for itself, and others acting on its behalf, a paid-up nonexclusive, irrevocable world-wide license in said article to reproduce, prepare derivative works, distribute copies to the public, and perform publicly and display publicly, by or on behalf of the Government. This work was supported by the U.S. Department of Energy, Office of Science under contract DE-AC02-06CH11357. The research work was funded by the Department of Energy Technology Commercialization Fund (TCF) project. The authors also acknowledge the computing resources provided on the "Blues" and "Bebop" high-performance computing clusters operated by the Laboratory Computing Resource Center (LCRC) at Argonne National Laboratory.